\documentclass[10pt,twocolumn,letterpaper]{article}
\usepackage[pagenumbers]{wacv} 
\usepackage[accsupp]{axessibility}

\usepackage{graphicx}
\usepackage{amsmath}
\usepackage{amssymb}
\usepackage{booktabs}
\usepackage{comment}
\usepackage{xcolor, soul}

\usepackage[
  headsep=3mm,
  top=3cm,
  bottom=2.3cm,
  left=1.8cm,
  right=2.4cm,
  head=30pt]{geometry}
\usepackage{fancyhdr}
\usepackage[pagebackref,breaklinks,colorlinks]{hyperref}

\usepackage[capitalize]{cleveref}
\crefname{section}{Sec.}{Secs.}
\Crefname{section}{Section}{Sections}
\Crefname{table}{Table}{Tables}
\crefname{table}{Tab.}{Tabs.}

\def\wacvPaperID{482} 
\def\confName{WACV}
\def\confYear{2024}

\begin{document}

\title{Rethinking Multimodal Content Moderation from an Asymmetric Angle with Mixed-modality}

\author{
Jialin Yuan $^\dagger$$^1$,
Ye Yu $^\dagger$$^2$,
 Gaurav Mittal $^2$,
 Matthew Hall $^2$,
 Sandra Sajeev $^2$,
 Mei Chen $^2$ \\
$^1$Oregon State University, $^2$Microsoft Inc. \\
{\tt \small \{yuanjial\}@oregonstate.edu}\\
{\tt \small \{yu.ye,gaurav.mittal,mei.chen,mathall,ssajeev\}@microsoft.com}
}

\maketitle

\newcommand\blfootnote[1]{%
  \begingroup
  \renewcommand\thefootnote{}\footnote{#1}%
  \addtocounter{footnote}{-1}%
  \endgroup
}
\blfootnote{$\dagger$ Equal contributions.}

\newcommand{\ourMethod}{AM3}
\newcommand{\ourMethodFullName}{Asymmetric Mixed-Modal Moderation}
\newcommand{\ourPreObjective}{\textsl{Cross-modality} Contrastive Loss}
\newcommand{\ourPreObjectiveSmall}{\textsl{cross-modality} contrastive loss}

\begin{abstract}
   There is a rapidly growing need for multimodal content moderation (CM) as more and more content on social media is multimodal in nature. Existing unimodal CM systems may fail to catch harmful content that crosses modalities (e.g., memes or videos), which may lead to severe consequences. In this paper, we present a novel CM model, {\ourMethodFullName} ({\ourMethod}), to target multimodal and unimodal CM tasks. Specifically, to address the asymmetry in semantics between vision and language, {\ourMethod} has a novel asymmetric fusion architecture that is designed to not only fuse the common knowledge in both modalities but also to exploit the unique information in each modality. Unlike previous works that focus on representing the two modalities into a similar feature space while overlooking the intrinsic difference between the information conveyed in multimodality and in unimodality (asymmetry in modalities), we propose a novel {\ourPreObjectiveSmall} to learn the unique knowledge that only appears in multimodality. This is critical as some harmful intent may only be conveyed through the intersection of both modalities. 
   With extensive experiments, we show that {\ourMethod} outperforms all existing state-of-the-art methods on both multimodal and unimodal CM benchmarks.
   
\end{abstract}

\section{Introduction} \label{sec:intro}
With the proliferation of multimodal social media and online gaming, user-generated content followed by recent AI-generated content (e.g., via DALL-E\cite{ramesh2022hierarchical}, GPT-3\cite{brown2020language}, ChatGPT\cite{van2023chatgpt} \etc) can spread across the internet at a faster rate than ever. While this enables free speech and facilitates information exchange, it comes with the risk of misuse for fake news \cite{nakamura2019r,vosoughi2018spread} and hate speech \cite{schmidt-wiegand-2017-survey,davidson2017automated}. 
\begin{figure}[t]
\centering
\includegraphics[width=1\linewidth]{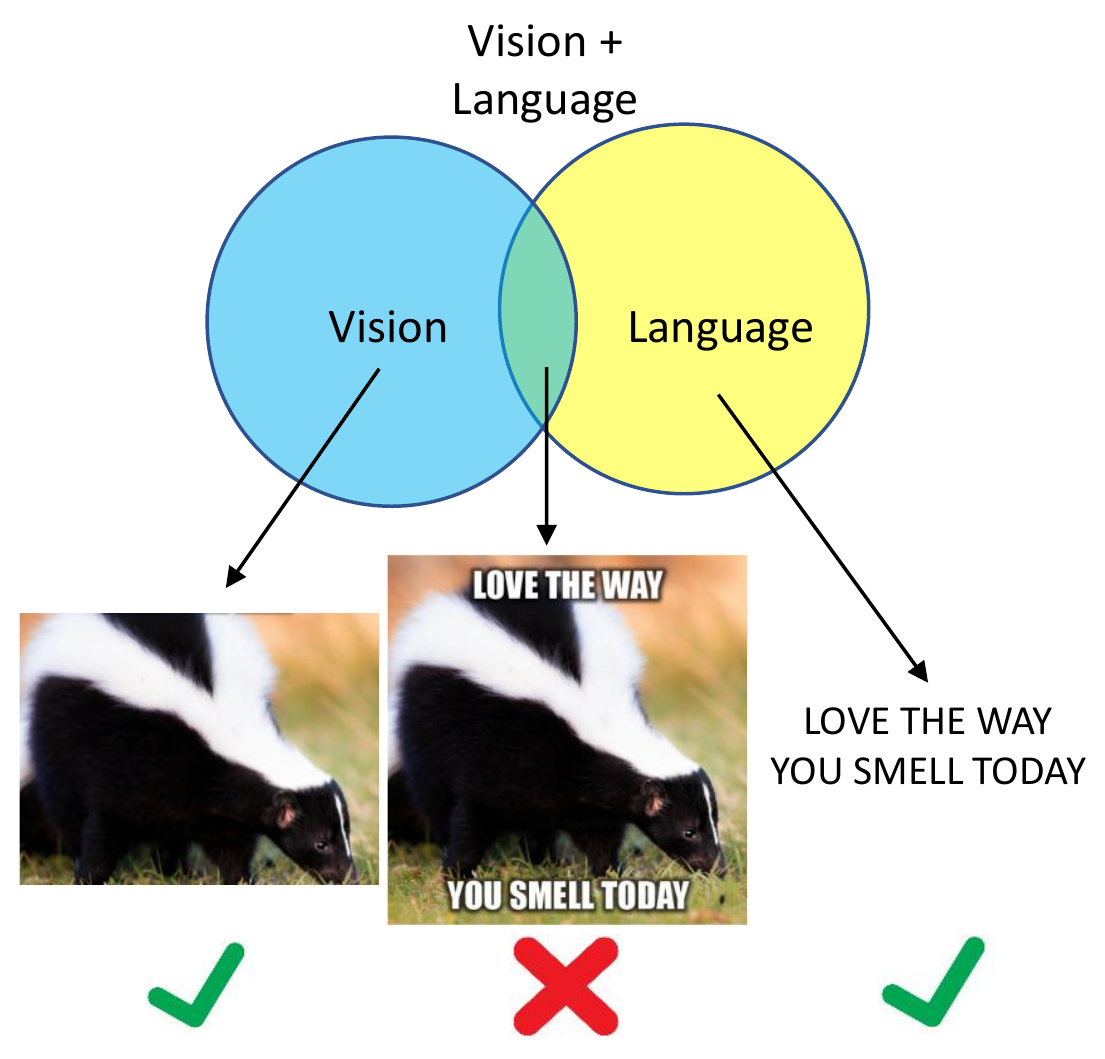}
\vspace{-0.5cm}
\caption{An example of a mean meme from Hateful Memes\cite{kiela2020hateful} for illustrative purposes. The unimodal vision and language are both benign while the multimodal meme is sarcastic and mean. This is not an actual example of the CM dataset $^\star$, which is hateful and would be distasteful to show here.}
\vspace{-0.5cm}
\label{fig:example}
\end{figure}

Leaving harmful content on social platforms can lead to harmful consequences, but moderating the tremendous amount of user/AI-generated content on the platforms manually is infeasible due to the large scale and can be harmful to the mental health of human moderators. Therefore, automated content moderation (CM) systems are necessary.
There has been extensive research on text-based content moderation \cite{vosoughi2018spread,waseem-etal-2017-understanding,waseem-hovy-2016-hateful}. Recently, there is a study on image-based pornographic content classification and sexual object detection tasks \cite{phanlspd}. As social platforms allow the use of different modalities, unsafe multimodal content may evade detection by existing unimodal content moderation systems. Hence, multimodal harmful content detection benchmarks \cite{kiela2020hateful,gomez2020exploring} have emerged followed by works \cite{zhu2020enhance,das2020detecting} aiming to automatically detect unsafe multimodal content, including child abuse material, violence, hate speech, sexual content, cyberbullying content, and disinformation \cite{banko-etal-2020-unified}.

One important form of multimodal content online is memes, which are a combination of image and short text. Understanding memes is a multimodal vision language (VL) task. As noted in previous studies \cite{gomez2020exploring}, offensive terms by themselves may not necessarily signify hate. It is the overall context that determines whether the intent is harmful or not. 
Fig.~\ref{fig:example} shows an example of mean meme, where the text by itself is just a compliment and the image 
also seems benign. However, when combining the two modalities the meme becomes sarcastic and mean. 
This example is for illustrative purpose only. For actual examples which are indeed hateful, please refer to supplementary.
To combat the spread of harmful VL content such as hateful memes on social platforms, different VL datasets have been constructed: Facebook proposed a Hateful Memes Challenge and constructed a corresponding dataset 
\cite{kiela2020hateful}, which contains memes designed to evade detection by unimodal methods. MMHS150K \cite{gomez2020exploring}, a large-scale image-text pair dataset originated from Twitter postings, is proposed to benchmark hate speech detection in multimodal publications.

In this work, we approach multimodal (image + text) harmful content detection and propose a novel mixed-modal (a mix of multimodality and unimodality) CM model, {\ourMethodFullName} (\ourMethod). 
Image and text are intrinsically different in the information they convey: text is more structured and semantically at a higher level  (usually describing the main components of an image while overlooking the subtle details, especially the background). On the other hand, image is unstructured: it is composed of pixels that can provide more low-level details of the context.
For example, an image caption is likely to focus on the foreground or the objects of interest in the image. It may contain semantic details like the color or shape of the objects, but unlikely to cover all the details, especially those in the background. We call this \textit{\textbf{asymmetry in semantics}} of VL content. 
To address this asymmetry, we propose a novel fusion transformer architecture that attempts to maintain the unique knowledge in each modality while fusing the information from the asymmetric semantic levels.
As shown in Fig.~\ref{fig:example}, the knowledge learned from the joint multimodality should contrast that from each unimodality due to this asymmetry in semantics. Sometimes this subtle missing part in unimodality is the determinant for content moderation decisions. We name the discrepancy in the information conveyed by multimodality and each unimodality \textit{\textbf{asymmetry in modalities}}. 
To tackle this challenge, we propose a novel contrastive loss between the representation learned from multimodality versus each unimodality. 
In order to learn domain-specific knowledge, we mix multimodal dataset with additional unimodal CM datasets in pretraining, similar to \cite{li-etal-2021-unimo}.
We call this \textit{\textbf{asymmetry in data}} as either modality may be missing in the data, so that the conventional multimodality (each sample contains both modalities) setup becomes mixed-modality (mix of multimodality and unimodality, where each sample may contain both modalities or each unimodality). 
By including unimodal CM dataset in pretraining, {\ourMethod} learns the domain-specific knowledge which helps the model adapt to the downstream tasks. Hence, the downstream CM task performance is improved.

We summarize the main contributions of work below,
\begin{itemize}
    \item \textbf{Asymmetry in semantics}: We propose a novel fusion transformer architecture to fuse different modalities asymmetrically. It enhances the unique knowledge in each modality while effectively fusing the information from the asymmetric semantic levels.
    \item \textbf{Asymmetry in modalities}: We design a novel contrastive loss to squeeze out the distinct knowledge that only exists in multimodality, which is essential in multimodal content moderation.
\end{itemize}

\section{Related Works} \label{sec:related}
\begin{figure*}[ht]
    \centering
    \includegraphics[width=\textwidth]{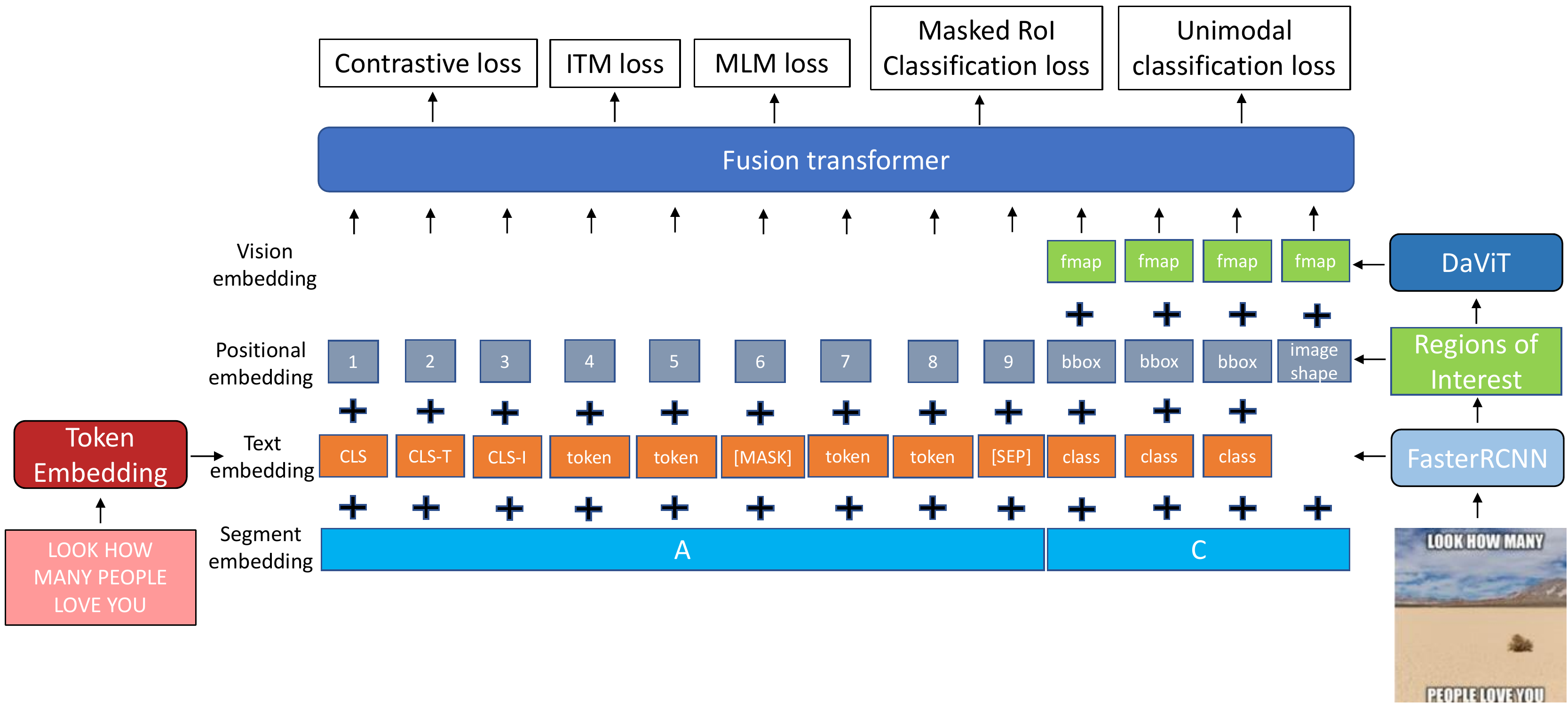}
    \caption{Architecture overview. It shows an example of the pretraining of {\ourMethod} with a $(T, I)$ input. For text inputs, we sum up text embeddings, positional embeddings, and the segment embeddings. Visual inputs consist of the text embeddings from detected objects' category labels, the feature map from the vision encoder, positional embeddings, and the segment embeddings. The positional embeddings of visual inputs are computed based on object bounding boxes so that they are permutation invariant to object order.}
    \label{fig:architecture}
\end{figure*}

\noindent\textbf{Harmful Content Detection.} As social media platforms have grown, so have the challenges of content moderation. These challenges have pushed platforms toward automated content moderation as a necessary tool for detecting harmful content.
Initially, most of the works are on text \cite{das2020detecting,baly-etal-2018-integrating}.
In \cite{davidson2017automated}, 25K Tweets are collected and annotated based on whether they contain hate speech keywords or have implicit hate. Logistic regression and SVM are tested to automatically detect hate speech. Besides web crawling data, a large scale machine generated dataset of toxic and benign text statements is provided in \cite{hartvigsen2022toxigen} using GPT-3 \cite{brown2020language}. These labels are then validated by human annotators, and over 95\% of the generated toxic labels are legitimately toxic. 
Over time, images and videos have gained more attention as visual contents are easier to consume and more popular to spread. A large scale dataset for pornographic visual content classification is given in \cite{phanlspd}. In \cite{soldner2019box}, videos of conversations are collected as a benchmark for deception detection.
Recently, multimodal harmful content detection has attracted more attention. Facebook proposed a Hateful Memes Challenge \cite{kiela2020hateful}, where each image is associated with a short text. The winner of the challenge \cite{zhu2020enhance} outperforms the other competitors significantly by leveraging external labels such as race, age, and entity. Following the same practice, DisMultiHate \cite{lee2021disentangling} further improves the performance by disentangling target entities in multimodal memes.  Hate-CLIPper\cite{kumar2022hate} proposes a method of intermediate fusion to alleviate the ambiguity alignment between image and text representations. The importance of each modality in the Hateful Memes dataset and the robustness of SOTA multimodal classfication algorithms are investigated in \cite{ma2022multimodal}. 


\noindent\textbf{Vision-Language Pretraining.} 
Recent years have witnessed rapid progress in vision-language pretraining (VLP) where vision and language modalities are jointly encoded using a fusion model. The success of BERT \cite{vaswani2017attention} inspired many follow-up multimodal fusion models, such as VL-bert\cite{su2019vl}, VinVL\cite{zhang2021vinvl}, SimVLM \cite{wang2021simvlm}, and OFA \cite{wang2022ofa}, where the text features are concatenated with vision features from image encoder and then fused by BERT or its variants. Besides the masked language modelling (MLM) loss used in BERT pretraining, various loss functions targeting multimodal feature fusion are used, e.g., image-text matching (ITM) loss, region-of-interest (RoI) classification loss. Most of these works learn the joint representation of vision and language through a symmetric feature encoding and fusion process. For example, VL-BERT\cite{su2019vl} constructs the multimodal inputs symmetrically where every multimodal feature map has the same components, i.e., text embedding, visual embedding, segment embedding, and positional embedding. Each text embedding is associated to the visual embedding of the entire image while each RoI visual embedding is associated with a dummy text embedding. This simple symmetric architecture enables the fusion of multiple modalities. However, each text token only contains a subset of the entire image. Linking the entire image embedding to it may introduce noise that decreases the performance. On the other hand, the dummy text embedding does not contain any meaningful information. VinVL\cite{zhang2021vinvl} simply concatenates text embeddings with the object label embeddings as well as RoI visual embeddings before feeding into the fusion transformer. It assumes that the text embeddings and the visual embeddings share the same (symmetric) level of knowledge and processes them equally.

Recent works on VL foundation models show that dual-encoder architectures can learn strong representation through contrastive objectives on large scale noisy image-text pairs \cite{radford2021learning,yuan2021florence,pham2021combined}. Florence \cite{yuan2021florence} developed a unified contrastive objective \cite{yang2022unified} in VLP that enables the model to be adapted for a wide range of vision and VL tasks. 
Flamingo \cite{alayrac2022flamingo} utilizes an 80B-parameter language model frozen in training and fused with a vision encoder. The huge capacity of Flamingo enables the state-of-the-art performance for few-shot learning.

Our method shares numerous ideas of the previous works mentioned above. However, we pivot to looking at the multimodal content moderation task from an asymmetric angle, both in architecture and data, and target mixed-modality (both multimodal and unimodal) downstream CM tasks. We exploit the discrepancy in vision, language, and multimodal VL pairs, to improve the model capability and training.

\section{Method} \label{sec:method}

In this paper, we present a novel fusion transformer architecture pretrained on both VL datasets and unimodality datasets. To tackle the asymmetry in semantics of CM VL content, we construct vision and language embeddings differently to encourage the model to capture essential knowledge in each modality. Meanwhile, we follow \cite{zhang2021vinvl} to utilize the object labels from detection as anchors to bridge the language with the corresponding image RoI features. Due to the asymmetry in modality, there is unique knowledge that only exists in the intersection of both modalities. To drive the model to obtain understanding of this, we introduce a novel contrastive loss, {\ourPreObjective}, 
as part of our pre-training tasks.
We use an asymmetric mix of multimodal datasets as well as domain-specific unimodal datasets in pretraining, where a domain-specific classification loss is included to improve downstream task performance.

\subsection{Model Architecture for Asymmetry in Semantics}
Fig.~\ref{fig:architecture} illustrates the overview architecture of {\ourMethod}. The model takes mixed modality input: $(T, I)$, $(T)$, or $(I)$, where $T$ represents the text if it exists, and $I$ is the image if it exists. Unlike previous works that try to unify the feature encoding process from both vision and language modalities, we construct the text inputs and visual inputs to the fusion transformer asymmetrically.
$T$ is first tokenized through a tokenizer and then fed to a token embedding layer whose outputs are added to positional embeddings and segment embeddings to generate the sequence of linguistic embeddings of text $\textbf{w}$. 
The image $I$ is processed as follows: we first use an object detection model 
to detect objects existing in the image. We also include a bounding box for the entire image (so the bounding box becomes the shape of the image) without an object category associated. For each object, its category label will go through the same token embedding layer as the text input to obtain its text embedding. Its bounding boxes are transferred to the positional embeddings of the RoI through a linear layer. This makes the positional embeddings permutation invariant to the input order of the objects. The visual feature of each RoI is encoded through a feature extractor. We then sum up the text embeddings of object labels, positional embeddings from object bounding boxes, the features from the RoIs, and segment embeddings to obtain the sequence of visual embeddings $\textbf{v}$. 
The concatenated pair of $(w, v)$ is fused through a fusion transformer.

\subsection{Cross-modality Contrastive Loss for Asymmetry in Modalities}\label{sec:3:objective}

\begin{figure*}
    \begin{subfigure}{.5\textwidth}
      \centering
      \includegraphics[width=.7\textwidth]{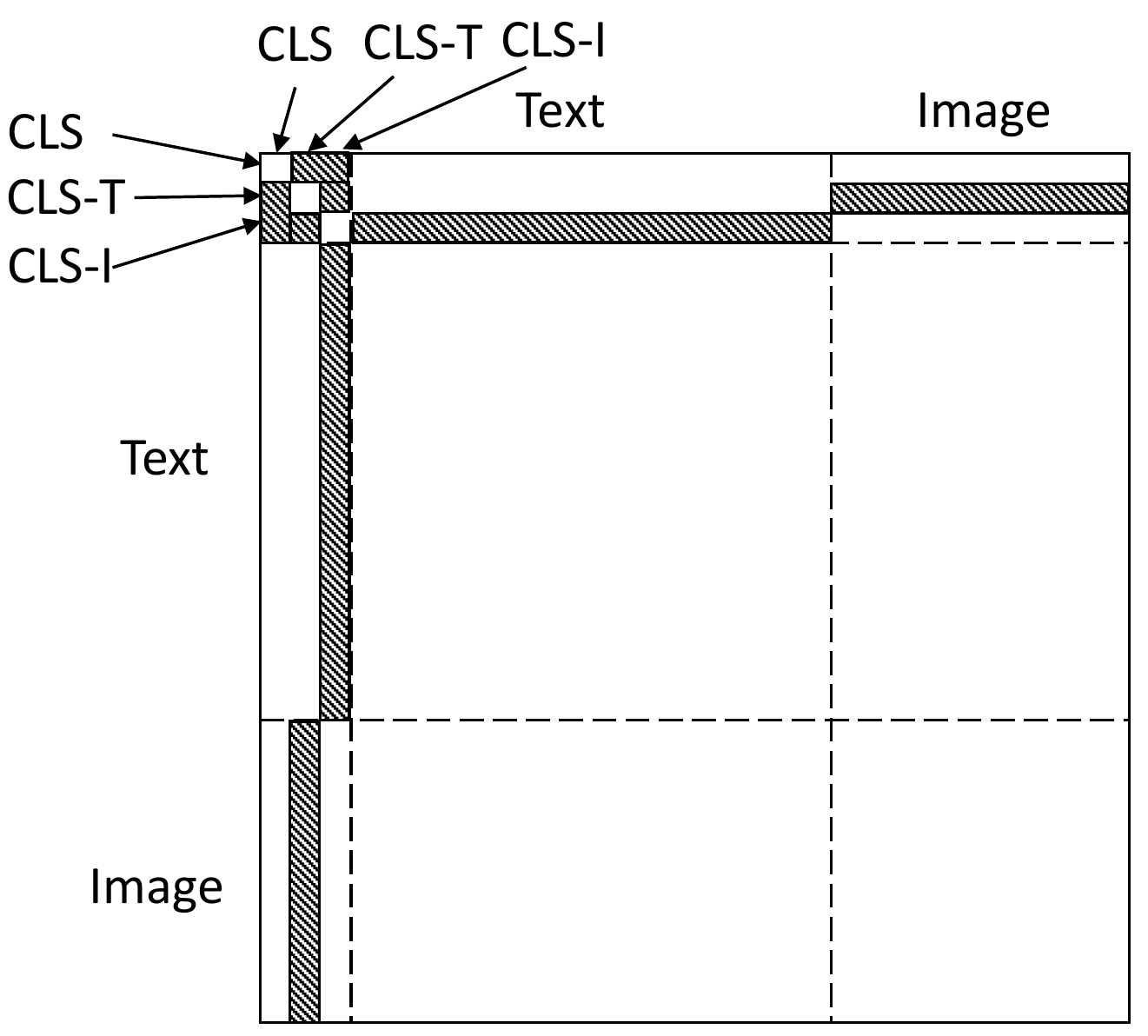}
      \caption{}
      \label{fig:mask}
    \end{subfigure}%
    \begin{subfigure}{.5\textwidth}
      \centering
      \includegraphics[width=.75\textwidth]{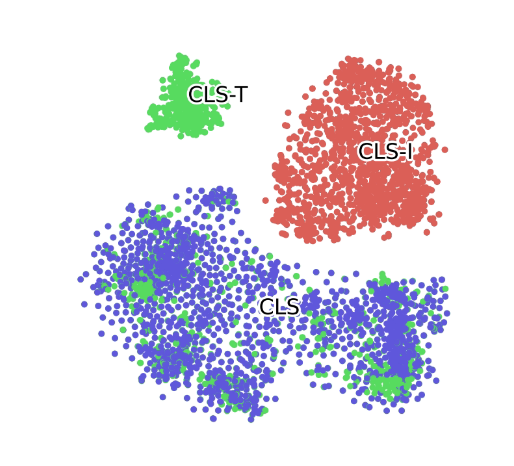}
      \caption{}
      \label{fig:visual-cls}
    \end{subfigure}
    \vspace{-0.7cm}
    \caption{(a): Modified attention mask for contrastive learning. Attention between image tokens (including CLS-I) and CLS-T tokens are masked out, and vice versa. (b): Visualization of the 3 CLS tokens from Hateful Memes after t-SNE\cite{gisbrecht2015parametric} reduction.}
    \vspace{-0.5cm}
\end{figure*}

\noindent\textbf{{\ourPreObjective}.} Due to the asymmetry in modalities, the capability of learning the unique knowledge only existing in the intersection of different modalities is critical to content moderation tasks, as demonstrated in Fig.~\ref{fig:example}. Therefore, we propose the {\ourPreObjectiveSmall}, as given in Equation(\ref{eq:contrastive-loss}):
\begin{equation}
    L_{con} = \max(0, \cos{(f_{VL}, f_{V})}) + \max(0, \cos{(f_{VL}, f_{L})})
    \label{eq:contrastive-loss}
\end{equation}
where $\cos$($\cdot$) is the cosine similarity function. $f_{VL}$, $f_{V}$, and $f_{L}$ are the CLS output tokens from the fusion transformer for image + text, image only, and text only, respectively. As shown in Fig.~\ref{fig:architecture}, 3 CLS tokens are added to the fusion transformer input. The CLS token is designed to summarize the multimodal knowledge from all tokens while the CLS-I and CLS-T tokens only extract information for vision and language tokens, respectively.
By summing up the similarity between $f_{VL}$ vs. $f_{V}$, and $f_{VL}$ vs. $f_{L}$ in the contrastive loss, we push the joint multimodal representation away from the unimodal representations, for the asymmetry in modality, forcing the model to learn the distinct semantic knowledge only in the intersection of both modalities. Fig.~\ref{fig:mask} illustrates how the attention mask is modified to compute the multimodal and unimodal representations: the attention between CLS and CLS-I/CLS-T as well as between CLS-I and CLS-T tokens are masked out to prevent information leak among different representations. The attention between the CLS-T token and all image tokens are also masked out, and vice versa. In this way, as displayed in Fig.~\ref{fig:visual-cls}, the CLS output token summarizes the fused information learned from both modalities while CLS-I and CLS-T only contain unimodal information in vision and language, respectively.

\noindent\textbf{Binary classification on domain-related datasets.} To help the model effectively adapt to the new domain 
(CM in our case) when porting a generic model to a specific domain, we include a domain-specific classification loss ($L_{domain}$) in our pretraining objectives.
We collect several content moderation related unimodality datasets discussed in Sec.\ref{subsec:de} into the pretraining corpus. When an input is from these datasets, its CLS output token is projected through a linear layer to predict if the input is harmful or not. We show that this domain-specific classification loss improves downstream performance on CM benchmarks. The domain-specific classification loss is:
\begin{equation}
    L_{domain} = -E_{f_{VL}}[\log P(c_d | f_{VL})]
    \label{eq:loss:domain}
\end{equation}
where $c_d$ is the domain category label and $f_{VL}$ is the fusion transformer output of the CLS token. In our experiment, we set $c_d = 1$ for harmful inputs while $c_d = 0$ for safe ones. For inputs from generic multimodal VL datasets, we set $c_d = -1$ so that they are ignored in the domain-specific classification task.

As shown in Fig.~\ref{fig:architecture}, there are $3$ additional pretraining objectives for multimodal fusion: the Masked Language Modeling loss ($L_{mlm}$ on the text tokens similar to \cite{huang2020pixel, zhang2021vinvl, su2019vl, kim2021vilt}, the Image-Text Maching loss ($L_{itm}$) which is computed on the CLS token of joint modalities same as \cite{huang2020pixel, kim2021vilt}, and the Masked RoI classification loss ($L_{roi{\text -}cls}$) similar to \cite{su2019vl}. Overall, our pretraining objective consists of terms as in Equation(\ref{eq:pretrain}):
\begin{equation}
    Loss = \alpha\ L_{con} + \beta\ L_{mlm} + \gamma\ L_{itm} + \lambda\ L_{roi{\text -}cls} + \omega\ L_{domain} 
    \label{eq:pretrain}
\end{equation}

where $\alpha$, $\beta$, $\gamma$, $\lambda$, and $\omega$ are coefficients to balance the various objectives. We set $\lambda$ to $0.2$ and all the other coefficients to $1$ throughout the experiments.

\section{Experiments} \label{sec:exp}
In this section, we first introduce the implementation details. We then discuss the results on downstream CM tasks. Finally, we show an ablation study on the proposed method.

\subsection{Implementation Details}\label{subsec:de}
\noindent\textbf{Pretraining.} As shown in Fig.~\ref{fig:architecture}, following \cite{li2020oscar}, we use pretrained FasterRCNN\cite{ren2015faster} for object detection, but other object detection models, like Yolo \cite{redmon2016you}, can be used as well. We use DaViT\cite{ding2022davit} as the vision encoder, which encodes the RoIs detected by FasterRCNN into vision embeddings. Both FasterRCNN and DaViT are frozen during training. 
We use $BERT_{base}$ ($Layers=12, Hidden\ size=768, Attention\ heads=12$) for text embedding and fusion transformer. 
The model is initialized with pretrained $BERT_{base}$ parameters and optimized using the AdamW optimizer with a base learning rate of $10^{-5}$ and weight decaying of $10^{-2}$. The learning rate was warmed up for 100 training steps and then decayed linearly to zero for the rest of the training. We use a probability of $0.15$ in MLM and Masked RoI classification random masking and $0.5$ in ITM random replacing. We assign segment tokens `C' to all visual features. For captions, we set segment tokens to `A', while for questions and answers, we use `A' and `B', respectively. We pretrain the model for $500K$ steps with a batch size of 6144 on 72 NVIDIA V100 GPUs.

\noindent\textbf{Pretraining corpus: } We construct our pretraining corpus based on three types of datasets: generic VL multimodal datasets, CM language datasets, and a CM vision dataset.

\begin{itemize}
    \item \textbf{Generic VL multimodal datasets.} 
    We build our corpus from image captioning and visual question-answer datasets, including COCO~\cite{lin2014microsoft}, Conceptual Captions (CC3M)~\cite{sharma2018conceptual}, SBU captions~\cite{ordonez2011im2text}, Flickr30k~\cite{young2014image}, CC12M~\cite{changpinyo2021conceptual}, Open-Images~\cite{kuznetsova2020open}, GQA~\cite{hudson2019gqa}, and VG-QAs datasets. Following~\cite{zhang2021vinvl}, machine-generated captions are used for Open-Images dataset, while captions and question-answer segments are used as text inputs for the other datasets.

    \item \textbf{CM language datasets.} 
    We use $4$ language datasets in CM domain: ToxiGen~\cite{hartvigsen2022toxigen}, Jigsaw~\cite{zaheri2020toxic}, HateXplain~\cite{mathew2021hatexplain}, and ImplicitHate~\cite{elsherief2021latent}, where we preprocess data so each sample has a harmful or safe label. We use \textsl{train} sets in pretraining for ToxiGen, Jigsaw, and HateXplain to avoid data leakage. For text samples without images, we pad [PAD] to vision embeddings. A text classification head predicts the label using the domain-specific classification objective.
    
    \item \textbf{CM vision dataset.} 
     We use LSPD~(Large-Scale Pornographic Dataset)~\cite{phanlspd} image dataset for CM vision task. Similar to the CM language datasets, we use the \textsl{train} set with binary annotation for pretraining and pad [PAD] tokens to the text inputs for fusion transformer. We use an image classification head to predict binary labels using the same classification objective.
\end{itemize}

\noindent\textbf{Downstream finetuning.} All the CM downstream tasks introduced in Sec.\ref{subsec:datasets} are formulated as classification tasks. The output token on CLS from the fusion transformer is fed into the classification head and trained with cross entropy loss. Hyperparameters including batch size, learning rate, and training epochs are searched for each task. All classification heads are implemented with an MLP consisting of $2$ linear layers and $1$ ReLU layer.

\noindent\textbf{Downstream inference.} In each task(Sec.~\ref{subsec:datasets}), we utilize the finetuning model and take the classification result from the CLS token as output, the model is named as \textbf{AM3}. On the downstream tasks, we conducted $5$ experiments with random seeds, reporting their mean and standard variation across the multiple finetuning models. On CM VL tasks, we assessed the model's ability in learning cross-modal knowledge using the best model with unimodal input as \textbf{AM3-text} and \textbf{AM3-image} respectively. Additionally, we combine the two unimodal results by taking the maximum classification probability as the predicted outcome, and refer to it as \textbf{AM3-max}.

\subsection{Downstream Datasets}\label{subsec:datasets}
To validate the effectiveness of {\ourMethod}, we adapt the pre-trained model over the content moderation tasks in different modalities.

For CM VL tasks, we adopted Hateful Memes, MMHS150K, and Fakeddit datasets. 
\begin{itemize}
    \item \textbf{Hateful Memes~\cite{kiela2020hateful}.} The Hateful Memes dataset 
    consists of more than 10,000 memes,
    some of which are specially designed so that the text phrases and images are benign when considered separately, but hateful when combined. Therefore, the typical unimodal methods cannot yield good performance on them.
    To compare with prior works\cite{zhu2020enhance,lee2021disentangling}, we use $2$ different setups: (1) we finetune our model on the \textsl{train} set and evaluate on the \textsl{dev seen} set. (2) We finetune our model on the combination of \textsl{train} and \textsl{dev unseen} sets and evaluate on the \textsl{test unseen} set. 
    The task uses the Area under Receiver Operating Characteristic curve (AUROC) and accuracy metrics. 
  
\item \textbf{MMHS150K~\cite{gomez2020exploring}} The MMHS150K dataset is based on Twitter data consisting of both image and text. 
    We perform binary classification to decide whether a sample is hate or non-hate. We finetune on the \textsl{train} and \textsl{val} sets and evaluate on the \textsl{test} set using F1-score, AUROC, and accuracy metrics.
    
\item \textbf{Fakeddit\cite{nakamura2019r}.} 
    The Fakeddit dataset 
    is a large-scale multimodal fake news dataset that consists of over $1$ million submissions from Reddit,
    a social news and discussion website where users can post submissions on various subreddits. 
    2-way, 3-way, and 6-way labels are provided for each sample. We follow the official dataset partition to only use multimodal samples.
    We focus on the 2-way classification and finetune our model on the \textsl{train} set. We compute accuracy on the \textsl{val} and \textsl{test} sets.
\end{itemize}

For CM text tasks, we use ToxiGen, HateXplain, and Jigsaw datasets.

\begin{itemize}
    \item \textbf{ToxiGen~\cite{hartvigsen2022toxigen}.} ToxiGen is a machine-generated dataset using the massive pretrained language model GPT-3\cite{brown2020language}. The dataset is designed to focus on creating hard-to-classify implicit abusive content in 13 minority groups. We use its \textsl{train} and \textsl{test} sets.
    The objective of the task is to predict if each sample is toxic or not and it is evaluated with AUROC. 

\item \textbf{HateXplain~\cite{mathew2021hatexplain}.} The HateXplain dataset is constructed by collecting posts from Twitter and Gab for research on Explainable Hate Speech Detection. 
The task is evaluated using AUROC, accuracy, and F1-score.

\item \textbf{Jigsaw~\cite{sahoo2022detecting}.}  The Jigsaw dataset is created using comments from Civil Comments
    for researchers to develop models to recognize toxicity and minimize this type of unintended bias with respect to mentions of identities, including gender, sexual orientation, and religious identity, \etc. We use the \textsl{train} and \textsl{test-public} splits for training and testing, respectively.
    AUROC is computed for evaluation.
\end{itemize}

We use LSPD for CM vision task.

\begin{itemize}
    \item \textbf{LSPD~\cite{phanlspd}.} 
    LSPD is constructed for visual pornography classification with $5$ categories: porn, hentai, drawing, sexy, and non-porn. 
    We followed the 
    porn/non-porn binary classification approach as \cite{phanlspd}, where the classes 'Hentai' and 'Porn' are grouped as 'porn', while all other classes were labeled 'non-porn' in the binary setting. To evaluate algorithms, 
    accuracy, precision, and recall are measured. 
\end{itemize}

\begin{table}[!htbp]
  \vspace{-0.2cm}
  \centering
  \caption{Comparisons to the state-of-the-art methods on Hateful Memes.}
  \vspace{-0.2cm}
  \resizebox{1\linewidth}{!}{%
    \begin{tabular}{l|cc|cc}
    \toprule
     & \multicolumn{2}{c}{Dev seen} & \multicolumn{2}{c}{Test unseen} \\
    Method & AUROC &  Accuracy & AUROC &  Accuracy \\
    \midrule
    ERNIE-VIL\cite{yu2021ernie}   & 78.7 & 69.0 & - & - \\
    Uniter\cite{chen2020uniter}  & 78.0  & 68.6 & 79.1 & 74.1  \\
    VILLA\cite{gan2020large}  & 78.5  & 71.2 & 80.0 & 75.1 \\
    VL-BERT\cite{su2019vl}   & 78.8 & 71.4 & 79.5 & 74.5  \\
    DisMultiHate\cite{lee2021disentangling}   & 82.8  & 75.8  & - & - \\ 
    \hline
     {\ourMethod}-text(Ours)  & 59.1 & 65.2 & 62.2 & 64.1  \\
     {\ourMethod}-image(Ours)  & 44.8 & 63.0 & 60.3 & 63.1  \\
     {\ourMethod}-max(Ours)  & 56.7 & 64.3 & 64.0 & 64.5  \\
     {\ourMethod}(Ours)  & \textbf{83.18($\pm$0.19)} & \textbf{75.98($\pm$0.67)} & \textbf{83.35 ($\pm$ 0.23)} & \textbf{76.95($\pm$0.36)}  \\
    \bottomrule
    \end{tabular}%
  }
  \vspace{-0.2cm}
  \label{tab:htfmm}%
\end{table}%

\begin{table}[!htbp]
  \centering
  \caption{Comparisons to the state-of-the-art methods on MMHS150K.}
  \vspace{-0.2cm}
  \resizebox{0.65\linewidth}{!}{%
    \begin{tabular}{lcc}
    \toprule
    Method & \multicolumn{1}{l}{AUROC} & \multicolumn{1}{l}{ Accuracy} \\
    \midrule
    TKM\cite{gomez2020exploring}   & 73.1  & 68.2  \\
    SCM\cite{gomez2020exploring}   & 73.2  & 68.5  \\
    FCM\cite{gomez2020exploring}   & 73.4  & 68.4  \\ \hline
    {\ourMethod}-text(Ours)  & 72.7 & 68.3 \\
    {\ourMethod}-image(Ours)  & 52.0 & 52.5  \\
    {\ourMethod}-max(Ours)  & 72.2 & 67.7  \\
    {\ourMethod}(Ours)  & \textbf{74.2 ($\pm$0.09)} & \textbf{68.57($\pm$0.79)}  \\
    \bottomrule
    \end{tabular}%
  }
  \vspace{-0.3cm}
  \label{tab:mmhs}%
\end{table}%

\begin{table}[!htbp]
  \centering
  \caption{Comparisons to the state-of-the-art methods on Fakeddit.}
  \vspace{-0.2cm}
  \resizebox{0.7\linewidth}{!}{%
    \begin{tabular}{lcc}
    \toprule
    Method  &  \multicolumn{1}{l}{Val acc.} & \multicolumn{1}{l}{Test acc.} \\
    \midrule
    BERT+ResNet50\cite{nakamura2019r}  & 89.3 & 89.1  \\
    MVAE+\cite{li2021entity} & - & 90.1  \\
    MDID\cite{kirchknopf2021multimodal} & 90.8 & 91.0  \\
    EMAF\cite{li2021entity} & - & 92.3 \\\hline
    {\ourMethod}-text(Ours)  & 82.23 & 82.41 \\
    {\ourMethod}-image(Ours)  & 76.2 & 75.9  \\
    {\ourMethod}-max(Ours)  & 83.1 & 83.3  \\
    {\ourMethod}(Ours)  & \textbf{93.04($\pm$0.21)} & \textbf{93.2($\pm$ 0.11)}  \\
    \bottomrule
    \end{tabular}%
  }
  \vspace{-0.2cm}
  \label{tab:fakeddit}%
\end{table}%

\subsection{Result Analysis}

\noindent\textbf{Performance comparison on VL tasks:} 
(1) Results on the Hateful Memes comparing to the state-of-the-art approaches are shown in Table \ref{tab:htfmm}.
We compare to the challenge winner's solutions discussed in \cite{zhu2020enhance}: ERNIE-Vil\cite{yu2021ernie}, UNITER\cite{chen2020uniter}, VILLA\cite{gan2020large}, and VL-BERT\cite{su2019vl}, where the results are reproduced in \cite{lee2021disentangling}. We also compare to the state-of-the-art solution, DisMultiHate\cite{lee2021disentangling}.
Furthermore, comparing to the baselines: {\ourMethod}-text, {\ourMethod}-image, and {\ourMethod}-max, the {\ourMethod} is at least $30\%$ better in terms of AUROC score, which verifies the efficiency of cross-modality understanding in the finetuned model.
We adopt the same data augmentation method proposed in \cite{zhu2020enhance}: we use Google Vision Web Entity Detection\cite{google2021webentity} to generate entity tags of each image used as part of the text input.
(2) The MMHS150K result is shown in Table \ref{tab:mmhs}. We compare to the Feature Concatenation Model (FCM), the Spatial Concatenation Model (SCM), and Texual Kernels Model (TKM) discussed in \cite{gomez2020exploring}, where they are all CNN + RNN models.
It is worth noting that the {\ourMethod}-text achieves a $72.7$ AUROC score, indicating that the dataset predominantly relies on its text modality. Merely considering the maximum classification probability to combine the image modality results leads to a decrease in performance.
(3) The Fakeddit result is shown in Table \ref{tab:fakeddit}. In \cite{nakamura2019r}, the authors of the Fakeddit dataset utilize BERT and ResNet50 to encode language and vision, respectively, and then use max-pooling to fuse the multimodal features. MAVE\cite{khattar2019mvae} is enhanced with BERT in \cite{li2021entity}, which is denoted as MAVE+ in the table. 
EMAF\cite{li2021entity} is set up with $BERT_{large\_uncased}$ $(Layers = 24, Hidden\ size = 1024, Attention\ heads = 16)$, which is computationally more expensive than our method.
The {\ourMethod} outperforms unimodal results, {\ourMethod}-text, {\ourMethod}-image, and {\ourMethod}-max,by at least $12.0\%$ in terms of accuracy.
On all three datasets, our proposed method achieves the best performance against previous state-of-the-art works. This demonstrates the efficacy of the proposed asymmetric mixed-modal approach. It effectively captures the distinct information that only appears in the intersection of modalities, which is critical in CM decision-making.

\noindent\textbf{Performance comparison on text tasks:} 
{\ourMethod} can also handle unimodal CM tasks.
We first evaluate our approach on the content moderation text tasks. (1) Results on ToxiGen Classification are listed in Table \ref{tab:toxigen}, where we compare to HateBERT\cite{caselli2020hatebert} and ToxDectRoBERTa\cite{zhou2021challenges} on the top-k only version of the dataset.
(2) Results on HateXplain dataset
are shown in Table \ref{tab:hatexplain}. Adaptive Length
Reduction (AdapLeR)\cite{modarressi2022adapler} is a method based on BERT while optimizing inference speed. BERT, BERT-HateXplain, BERT-MLM, BERT-RP, and BERT-MRP are different BERT variants discussed in \cite{kim2022hate}.
(3) Results on Jigsaw are shown in Table \ref{tab:jigsaw}, where we compare to the Toxiciology\cite{jigsawcompetition} and Limerobot\cite{jigsawcompetition}, the top 2 solutions on the leaderboard.
On all three CM text datasets, our approach outperforms all the state-of-the-art language models, suggesting the efficacy of the proposed method on mixed-modal (both multimodal and unimodal) downstream CM tasks.

\begin{table}[!htbp]
  \centering
  \caption{Comparisons to the state-of-the-art methods on ToxiGen.}
  \vspace{-0.2cm}
  \resizebox{0.6\linewidth}{!}{%
    \begin{tabular}{lc}
    \toprule
    Method & \multicolumn{1}{l}{AUROC} \\
    \midrule
    ToxDectRoBERTa\cite{hartvigsen2022toxigen}   & 85.0  \\
    HateBERT\cite{hartvigsen2022toxigen}   & 88.0  \\ \hline
    {\ourMethod}(Ours)  & \textbf{91.52 ($\pm$ 0.16)}  \\
    \bottomrule
    \end{tabular}%
    }
  \vspace{-0.5cm}
  \label{tab:toxigen}%
\end{table}%

\begin{table}[!htbp]
  \centering
  \caption{Comparisons to the state-of-the-art methods on HateXplain.}
  \vspace{-0.1cm}
  \resizebox{0.9\linewidth}{!}{%
    \begin{tabular}{lccc}
    \toprule
    Method & \multicolumn{1}{l}{AUROC} & \multicolumn{1}{l}{Accuracy} & \multicolumn{1}{l}{F1} \\
    \midrule
    AdaptLeR\cite{modarressi2022adapler}  &- & 68.6  & -  \\
    BERT\cite{subramaniam2022exploring}  &85.1 & 68.9  & 68.2  \\
    BERT-HateXplain\cite{mathew2021hatexplain}  &85.1 & 69.8  & 68.7  \\
    BERT-MLM\cite{kim2022hate}  &85.4 & 70.0  & 67.5  \\
    BERT-RP\cite{kim2022hate}  &85.3 & 70.7  & 69.3  \\
    BERT-MRP\cite{kim2022hate}  &86.2 & 70.4  & 69.9  \\\hline
    {\ourMethod}(Ours)  & \textbf{88.25 ($\pm$0.25)} & \textbf{81.17($\pm$0.45)} & \textbf{80.37($\pm$0.42)}  \\
    \bottomrule
    \end{tabular}%
  }
  \vspace{-0.3cm}
  \label{tab:hatexplain}%
\end{table}%

\begin{table}[!htbp]
  \centering
  \caption{Comparisons to the state-of-the-art methods on Jigsaw.}
  \vspace{-0.3cm}
  \resizebox{0.5\linewidth}{!}{%
    \begin{tabular}{lc}
    \toprule
    Method & \multicolumn{1}{l}{AUROC} \\
    \midrule
    Limerobot\cite{jigsawcompetition}   & 94.7  \\
    Toxiciology\cite{jigsawcompetition}   & 94.7  \\\hline
    {\ourMethod}(Ours)  & \textbf{95.76($\pm$0.27)}  \\
    \bottomrule
    \end{tabular}%
    }
  \vspace{-0.3cm}
  \label{tab:jigsaw}%
\end{table}%

\noindent\textbf{Performance comparison on vision task:} 
Results on the LSPD dataset are presented in Table 
\ref{tab:lspd-binary}, where we compare them to the outcomes of different methods discussed in \cite{phanlspd}. 
Our approach outperforms previous state-of-the-art methods in terms of accuracy and obtained the highest recall score. Similar to the CM text datasets, this shows the capability of our mixed-modal method on downstream CM vision task, benefiting from a richer representation space with the mixed-modality pretraining. 


\begin{table*}[!ht]
  \centering
  \caption{Ablation study of mixed-modality and {\ourPreObjectiveSmall}.}
  \vspace{-0.2cm}
  \resizebox{0.85\textwidth}{!}{%
    \begin{tabular}{ccc|cc|cc}
    \toprule
     & & & \multicolumn{2}{c}{Hateful Memes} & \multicolumn{2}{|c}{MMHS150K} \\
    Text dataset & Vision dataset & {\ourPreObjective} &  AUROC & Accuracy & AUROC & Accuracy \\
    \midrule
   no & no & no & 80.28($\pm$0.18) & 74.28($\pm$0.29) & 71.91 ($\pm$ 0.02) & 67.44($\pm$0.05) \\ 
   no & no & yes & 81.18($\pm$1.0) & 74.82($\pm$0.51) & 72.76($\pm$0.08) & 68.21($\pm$0.05)  \\
   no & yes & no & 80.65($\pm$0.2) & 73.82($\pm$0.36) & 72.52($\pm$0.05) & 68.03($\pm$0.09) \\ 
   no & yes & yes & 81.49($\pm$1.25) & 74.98($\pm$0.46) & 72.98($\pm$0.13) & 68.18($\pm$0.05) \\
   yes & no & no & 82.39($\pm$0.08) & 76.38($\pm$0.02) & 72.79($\pm$0.21) & 68.70($\pm$0.09) \\
   yes & no & yes & 82.65($\pm$0.2) & 75.92($\pm$0.17) & 73.36($\pm$0.07) & 68.62($\pm$0.09)  \\
   yes & yes & no & 82.44($\pm$0.23) & 75.5($\pm$0.31) & 72.85($\pm$0.10) & 68.45($\pm$0.07) \\
   yes & yes & yes & \textbf{82.94($\pm$0.15)} & \textbf{76.45($\pm$0.36)} & \textbf{73.96($\pm$0.08)} & \textbf{68.73($\pm$0.05)} \\
    \bottomrule
    \end{tabular}%
  }
  \vspace{-0.2cm}
  \label{tab:ablation:corpus}%
  
\end{table*}%


\begin{table}[!htbp]
  \centering
  \caption{Comparisons to state-of-the-art methods on LSPD for binary classification.}
  \vspace{-0.2cm}
  \resizebox{0.93\linewidth}{!}{%
  \begin{tabular}{l|ccc}
    \toprule
    Method & \multicolumn{1}{l}{Accuray} & \multicolumn{1}{l}{Precision} & \multicolumn{1}{l}{Recall} \\
    \midrule
    Mask-RCNN  & 86.70 & \textbf{98.33} & 88.00  \\
    YOLOv4     & 92.59 & 97.03 & 87.86  \\
    SSD        & 85.32 & 94.11  & 85.64  \\
    Cascaded Mask RCNN  &92.62 & 95.01 & 89.95  \\
    CNN classifier  & 87.22 & 84.86 & 90.59  \\
    \hline
    {\ourMethod}(Ours) & \textbf{ 92.86($\pm$0.03)} & 92.85 ($\pm$0.15) & \textbf{92.73($\pm$0.14)}  \\
    \bottomrule
  \end{tabular} 
  }
  \vspace{-0.1cm}
  \label{tab:lspd-binary}%
\end{table}%

\begin{table}[!htbp]
  \centering
  \caption{Ablation study of fusion architecture design.}
  \vspace{-0.2cm}
  \resizebox{1.0\linewidth}{!}{%
    \begin{tabular}{c|cc|cc}
    \toprule
     & \multicolumn{2}{c}{Hateful Memes} & \multicolumn{2}{|c}{MMHS150K} \\
    Architecture & AUROC & Accuracy & AUROC & Accuracy \\
    \midrule
    $arch_{VL-Bert}$  & 80.26($\pm$0.38) & 75.72($\pm$0.62) & 73.31($\pm$0.15) & 68.54($\pm$0.06) \\ 
    $arch_{vinVL}$ & 80.53($\pm$0.23) & 75.49($\pm$0.17) & 73.39($\pm$0.12) & 68.4($\pm$0.07)  \\
    $arch_{bbox-position}$ & 80.74($\pm$0.29) & 75.0($\pm$0.09) & 73.43($\pm$0.21) & 68.4($\pm$0.11) \\
    {\ourMethod}(Ours) & \textbf{82.94($\pm$0.15)} & \textbf{76.45($\pm$0.36)} & \textbf{73.96($\pm$0.08)} & \textbf{68.73($\pm$0.05)} \\
    \bottomrule
    \end{tabular}%
  }
  \vspace{-0.5cm}
  \label{tab:ablation:modelArch}%
\end{table}%

\subsection{Ablation Study}\label{sub:ab}

We selected Hateful Memes and MMHS150K for the ablation study of different design choices. To accelerate the analysis, all ablations are performed on a smaller pretraining corpus (Flickr30k, SBU, and COCO), and we pretrain our model for $50K$ iterations.

\noindent\textbf{Model Architecture.} To understand the effect of our proposed asymmetric fusion transformer, we create two fusion transformer variants following VL-Bert ($arch_{VL-Bert}$ \cite{su2019vl}) and vinVL ($arch_{vinVL}$ \cite{zhang2021vinvl}), two symmetric fusion designs. Specifically, $arch_{VL-Bert}$ constructs the multimodal embeddings symmetrically so that each text embedding adds to the visual feature of the entire image while each RoI visual embedding adds to a text embedding of a dummy token. $arch_{vinVL}$ creates multimodal embeddings for fusion transformer by simply concatenating the text embeddings from text input and object detection labels, along with visual embeddings. As shown in Table \ref{tab:ablation:modelArch}, our proposed asymmetric fusion architecture outperforms both symmetric designs, indicating the efficacy of our asymmetric fusion architecture in response to the \textit{asymmetry in semantics}.

\noindent\textbf{Vision Position Embedding from Bounding Box.} To validate the effectiveness of using bounding boxes for positional embeddings, we created a variant using the counting index of the tokens for positional embeddings (used in \cite{su2019vl,zhang2021vinvl}). As shown in Table \ref{tab:ablation:modelArch}, positional embeddings generated from bounding box captures the ordering information in image (permutation invariant to the input order). Therefore, it achieves a better performance.

\noindent\textbf{\ourPreObjective.} As shown in Table \ref{tab:ablation:corpus}, the average baseline scores on Hateful Memes and MMHS150K are $80.28$\% and $71.91$\%, respectively, measured in terms of AUROC. By adopting the {\ourPreObjectiveSmall}, the score is improved by $+1.1$\% on Hateful Memes and $+1.2$\% on MMHS150K. The significant improvements show that our approach to the \textsl{asymmetry in modalities} has a strong capability to capture distinct knowledge from the intersection of different modalities.

\noindent\textbf{Pretraining on Unimodal CM Datasets.} Table \ref{tab:ablation:corpus} shows the result $w/$ and $w/o$ the unimodal CM datasets in the pretraining corpus. Using the CM text datasets improves the task scores by $+2.6$\% and $+1.2$\% from baseline, respectively. Using the CM image datasets improves the score by $+0.5$\% and $0.8\%$, respectively. This shows that introducing \textsl{asymmetry in data} into the pretraining stage, with the datasets relevant to the domain, is effective and can improve downstream tasks by a significant margin. 

\noindent\textbf{Combination of {\ourPreObjective} and Unimodal CM Datasets.} As shown in Table \ref{tab:ablation:corpus}, utilizing the CM text dataset and the CM vision dataset together leads to further improvement ($+0.6$\% on Hateful Memes and $+0.8$\% on MMHS150K) in comparison to the best score when using either CM text dataset or CM vision dataset. Adding {\ourPreObjectiveSmall} on top of the unimodal CM text and vision datasets further improve the performance: when enabling all of these components, we achieve the highest average AUROC score of $82.94$\% for Hateful Memes and $73.96$\% for MMHS150K. It indicates the efficacy of our proposed method.

\section{Conclusion}
In this paper, we present a novel mixed-modal CM model, {\ourMethodFullName} ({\ourMethod}), for both multimodal and unimodal content moderation. We propose an asymmetric fusion architecture to fuse multimodal knowledge. 
Furthermore, we design a novel {\ourPreObjectiveSmall} to learn the distinct knowledge that can only be conveyed when combining both modalities, which is critical for multimodal CM tasks. Besides using multimodal VL datasets, we also include unimodal CM datasets in pretraining, which not only relaxes data constraints but also improves downstream task performance. With extensive experiments, we show {\ourMethod} achieves the new state-of-the art on various multimodal and unimodal CM benchmarks.

{\small
\bibliographystyle{ieee_fullname}
\bibliography{egbib}
}

\clearpage
\section{Supplementary Material}
In this supplementary material, additional examples from the Hateful Memes dataset are presented. 

\subsection{Successful examples}
\begin{figure*}[t]
\centering
\includegraphics[width=0.8\linewidth]{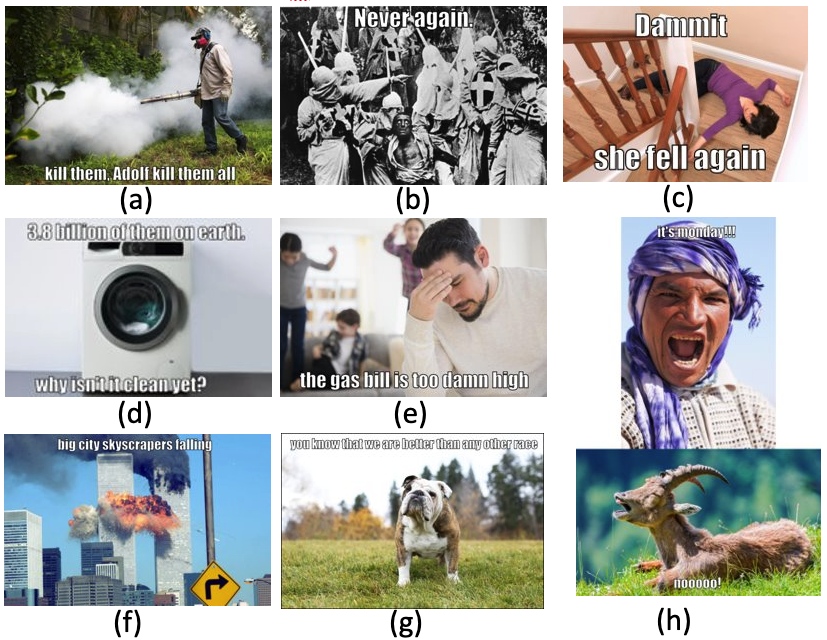}
\caption{\textcolor{red}{[CONTENT WARNING]} non-hate examples on multi-modalities that {\ourMethod} correctly detected.}
\vspace{-0.1cm}
\label{fig:visual-succ-nohate-mix}
\end{figure*}

\begin{figure*}[!htp]
\centering
\includegraphics[width=0.8\linewidth]{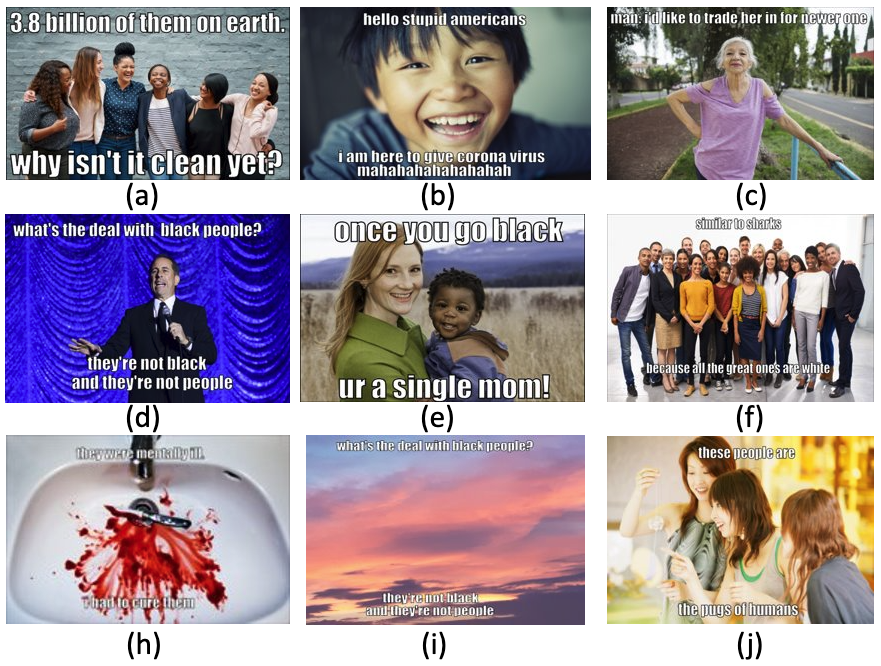}
\caption{\textcolor{red}{[CONTENT WARNING]} Hateful examples on uni-modality that {\ourMethod} correctly detected.}
\vspace{-0.6cm}
\label{fig:visual-succ-hate-single}
\end{figure*}

\begin{figure*}[!htp]
\centering
\includegraphics[width=0.95\linewidth]{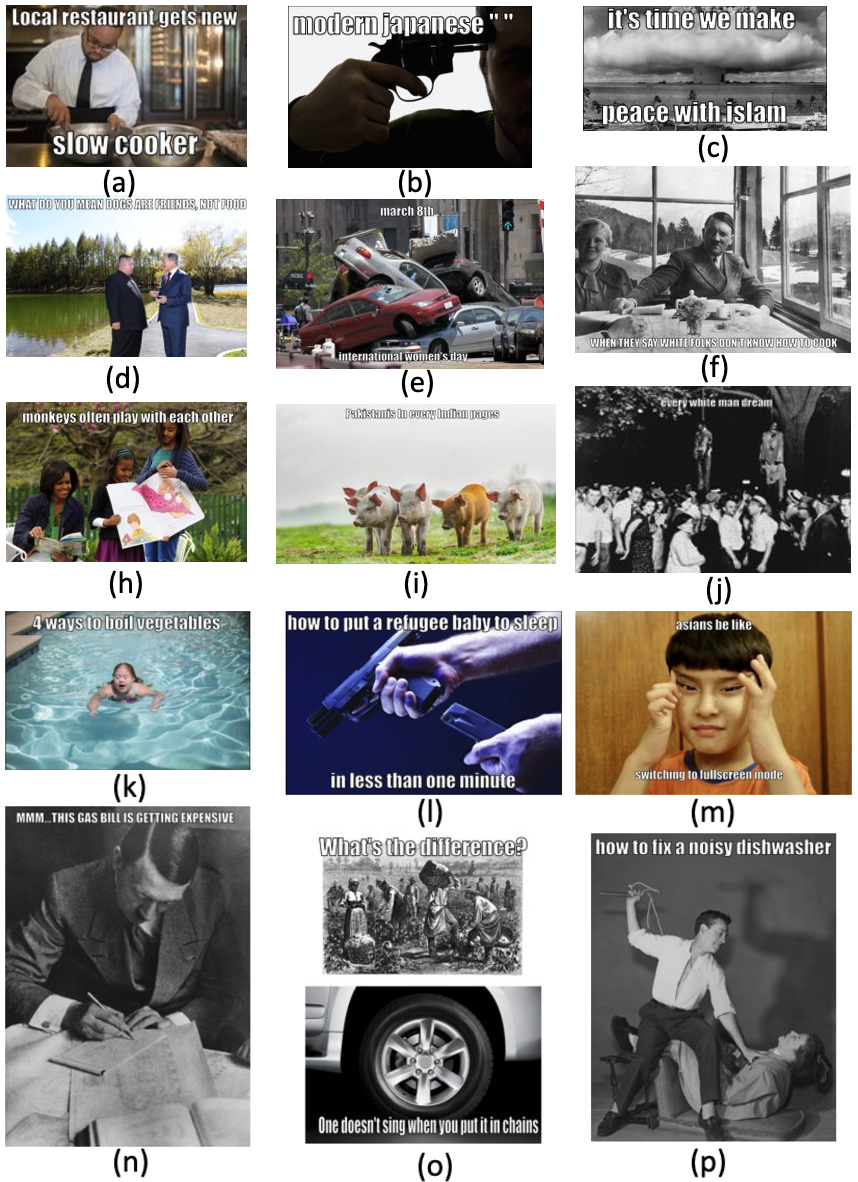}
\caption{\textcolor{red}{[CONTENT WARNING]} Hateful examples on multi-modalities that {\ourMethod} correctly detected.}
\vspace{-0.6cm}
\label{fig:visual-succ-hate-mix}
\end{figure*}

We show $3$ groups of successful examples in Fig. \ref{fig:visual-succ-nohate-mix}, \ref{fig:visual-succ-hate-single}, and \ref{fig:visual-succ-hate-mix}.
\begin{itemize}
    \item Fig.\ref{fig:visual-succ-nohate-mix} presents \textsl{examples that are not offensive in multimodality but are hateful on unimodality}. 
    \item Fig.\ref{fig:visual-succ-hate-single} shows \textsl{the correctly detected offensive examples that are offensive in unimodality}.
    \item Fig.\ref{fig:visual-succ-hate-mix} displays \textsl{the correctly detected offensive examples that are only offensive in multimodality}. 
\end{itemize}

The examples indicate that accurately predicting hate or non-hate is more complex than just mapping the two modalities. 
For instance, the text in Fig.\ref{fig:visual-succ-hate-mix}(n) and Fig.\ref{fig:visual-succ-nohate-mix}(e) conveys similar meanings, one is \textbf{hateful} and the other is \textbf{non-hate} when considering the objects in the images. Meanwhile, {\ourMethod} does not simply learn \textbf{Adolf (Hitler)} as \textbf{hateful} as shown in Fig.\ref{fig:visual-succ-nohate-mix}(a). 
The image in Fig.\ref{fig:visual-succ-nohate-mix}(h) and the Fig.\textcolor{red}{4}(a) in the main paper are similar and their text are both non-offensive, with the  understanding of multi-modalities in an asymmetric manner, {\ourMethod} successfully predict them as \textbf{non-hate} and \textbf{hateful}.
The text in Fig.\ref{fig:visual-succ-nohate-mix}(c), (d) and (e) seem to be \textbf{hateful}, with accurate detection and understanding of the objects in the image, the examples turn to be \textbf{non-hate}.

\subsection{Failure examples}
\begin{figure*}[!htp]
\centering
\includegraphics[width=0.95\linewidth]{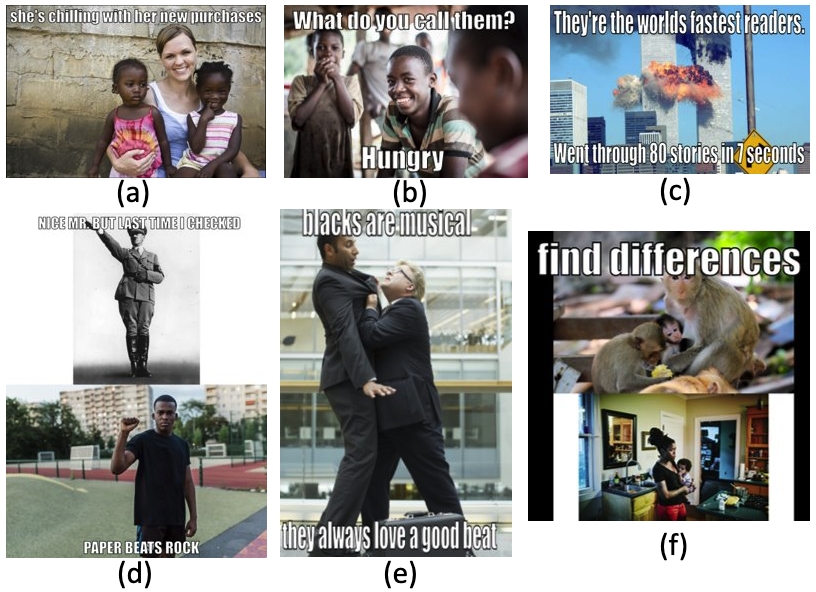}
\caption{\textcolor{red}{[CONTENT WARNING]} Hateful examples on multi-modalities that {\ourMethod} failed to detect.}
\vspace{-0.6cm}
\label{fig:visual-fail-hate-mix}
\end{figure*}

We provide some \textsl{offensive} examples in Fig.\ref{fig:visual-fail-hate-mix} that AM3 failed to detect.
In (a), (b), and (c), the false prediction is because of missing of the extended attributes, 'adoption,' 'poor,' and '9.11 attack'.
In (d), detecting the gestures 'paper' and 'rock' is necessary for correct prediction. However, {\ourMethod} does not support detection of gestures, leading to a false prediction.
In (e) and (f), understanding the relationship of the objects in the image is crucial to classify them as \textbf{hateful}. Improving the method to address these three problems may further enhance the task.

\end{document}